\def\year{2021}\relax
\documentclass[letterpaper]{article} % DO NOT CHANGE THIS
\usepackage{aaai21}  % DO NOT CHANGE THIS
\usepackage{times}  % DO NOT CHANGE THIS
\usepackage{helvet} % DO NOT CHANGE THIS
\usepackage{courier}  % DO NOT CHANGE THIS
\usepackage[hyphens]{url}  % DO NOT CHANGE THIS
\usepackage{graphicx} % DO NOT CHANGE THIS
\usepackage{natbib}  % DO NOT CHANGE THIS AND DO NOT ADD ANY OPTIONS TO IT
\usepackage{caption} % DO NOT CHANGE THIS AND DO NOT ADD ANY OPTIONS TO IT
\usepackage{bbm}
\usepackage{amsmath}
\usepackage{multirow}
\usepackage{amssymb}
\usepackage[switch]{lineno}

\makeatletter  
\newif\if@restonecol  
\makeatother

\usepackage[linesnumbered,ruled,vlined]{algorithm2e}%[ruled,vlined]{  
\usepackage{algpseudocode}  
\usepackage{amsmath}  
\usepackage{tabularx}
\usepackage{xcolor}
\usepackage{acronym}
\acrodef{DNN}{Deep Neural Networks}
\acrodef{CNN}{Convolutional Neural Networks}
\acrodef{SOTA}{State-of-the-Art}
\acrodef{CLOR}{Contour Landmark Offset Regression}
\acrodef{LHR}{Landmark Heatmap Regression}
\acrodef{MFE}{Motion Field Estimation}
\acrodef{NME}{Normalized Mean Error}
\acrodef{AUC}{Area Under the Curve}

\begin{document}
%\linenumbers
	
\title{Teacher-Student Asynchronous Learning with Multi-Source Consistency\\ for Facial Landmark Detection}
\author{
   Rongye Meng,
    Sanping Zhou,
    Xingyu Wan,
    Mengliu Li,
    Jinjun Wang
    \\
}
\affiliations{
    mengrongye@gmail.com, sanpinzhou@stu.xjtu.edu.cn,wanxingyu@stu.xjtu.edu.cn\\
    meng1996liu@stu.xjtu.edu.cn, jinjun@mail.xjtu.edu.cn
}

	\maketitle
	
	\begin{abstract}
		
		Due to high annotation cost of large-scale facial landmark detection tasks in videos, a semi-supervised paradigm that uses self-training for mining high-quality pseudo-labels to participate in training has been proposed by researchers. However, self-training based methods often train with an gradually increasing number of samples, whose performances vary a lot depending on the number of pseudo-labeled samples added. 
		
		In this paper, we propose a teacher-student asynchronous learning~(TSAL) framework based on multi-source supervision signal consistency criterion, which implicitly mines pseudo-labels through consistency constraints. Specifically, TSAL framework contains two models with exactly the same structure. The radical student uses multi-source supervision signals from the same task to update parameters, while the calm teacher uses single-source supervision signal to update parameters. In order to reasonably absorb student's suggestions, teacher's parameters are updated again through recursive average filtering. The experimental results prove that asynchronous-learning framework can effectively filter noise in multi-source supervision signals, thereby mining the pseudo-labels which are more significant for network parameter updating. And extensive experiments on 300W, AFLW and 300VW benchmarks show that TSAL framework achieves state-of-the-art performance.
		
	\end{abstract}

	%\begin{itemize}
	%\item You must use the 2021 AAAI Press \LaTeX{} style file and the aaai21.bst bibliography style files, which are located in the 2021 AAAI Author Kit (aaai21.sty, aaai21.bst).
	%\item You must complete, sign, and return by the deadline the AAAI copyright form (unless directed by AAAI Press to use the AAAI Distribution License instead).
	%\item You must read and format your paper source and PDF according to the formatting instructions for authors.
	%\item You must submit your electronic files and abstract using our electronic submission form \textbf{on time.}
	%\item You must pay any required page or formatting charges to AAAI Press so that they are received by the deadline.
	%\item You must check your paper before submitting it, ensuring that it compiles without error, and complies with the guidelines found in the AAAI Author Kit.
	%\end{itemize}

	\begin{figure}[t]
		\centering
		\includegraphics[width=1.0\columnwidth]{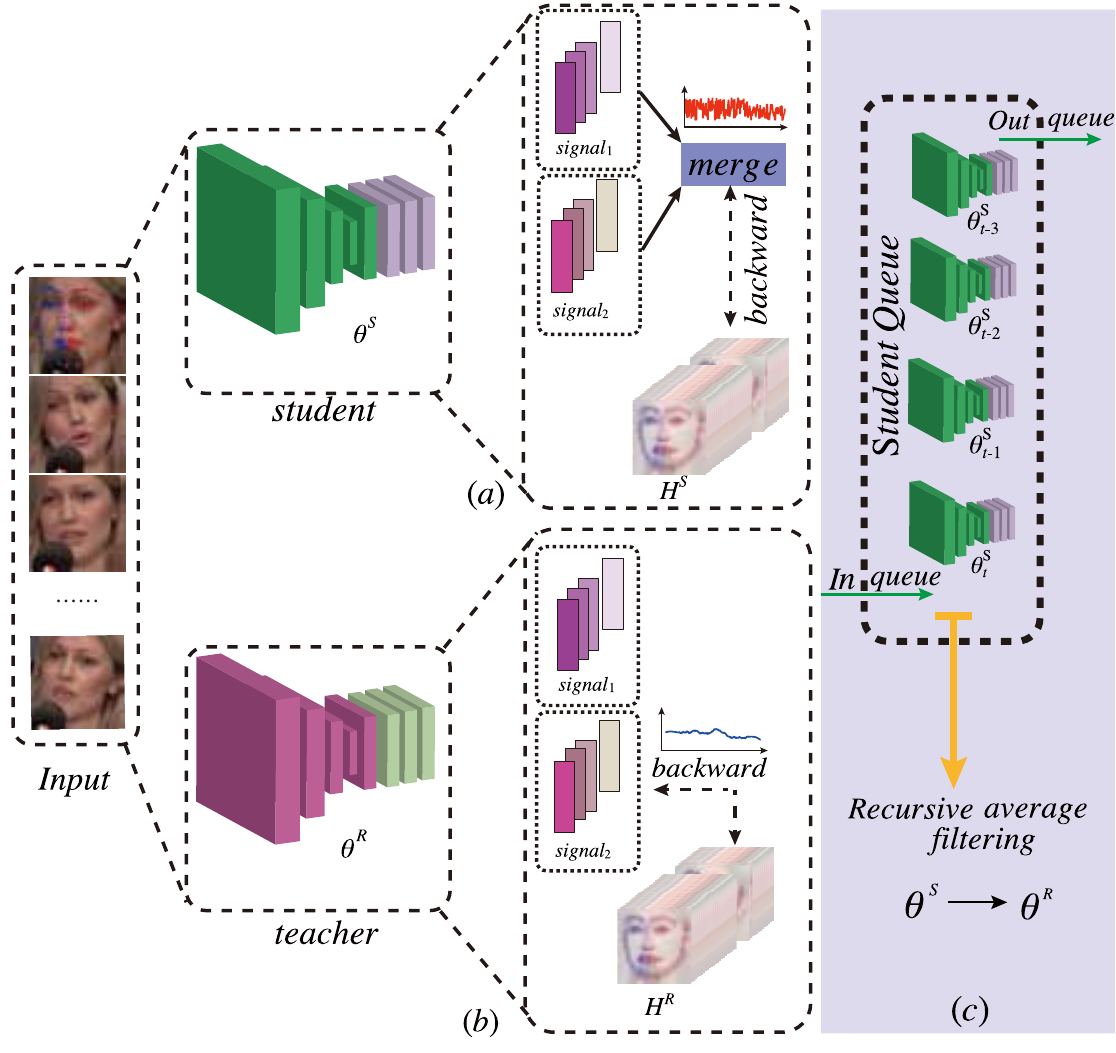} % Reduce the figure size so that it is slightly narrower than the column. Don't use precise values for figure width.This setup will avoid overfull boxes.
		%\vspace{-0.25cm}
	\caption{The asynchronous learning mechanism of teacher-student model. ~(a) Radical student is supervised by multi-source signals with greater disturbance.~(b) Calm teacher is supervised by single-source signal with less disturbance. ~(c) The second update of the teacher's parameters by recursive average filtering students in queue.}
		\label{fig1}
	\end{figure}
	
	\section{Introduction}
	Facial landmark detection is widely used as a preliminary task in face related computer vision application~\cite{Sun_2013_ICCV, Liu_2017_CVPR, Benitez-Quiroz_2016_CVPR, Li_2017_CVPR, 1227983, Dou_2017_CVPR, cite-key, HU2017366, Thies_2016_CVPR}. Although there are several public datasets with labeled facial landmark available~\cite{Sagonas_2013_ICCV_Workshops, 6130513, Tzimiropoulos_2015_CVPR, Shen_2015_ICCV_Workshops}, the process to mark the precise location of all the landmarks in large scale image or video collections is very time consuming, which renders fully-supervised training of \ac{DNN} based facial landmark detector tedious and costly~\cite{Dong_2019_ICCV}.
	
	\begin{figure*}[t]
		\centering
		\includegraphics[width=1.0\textwidth]{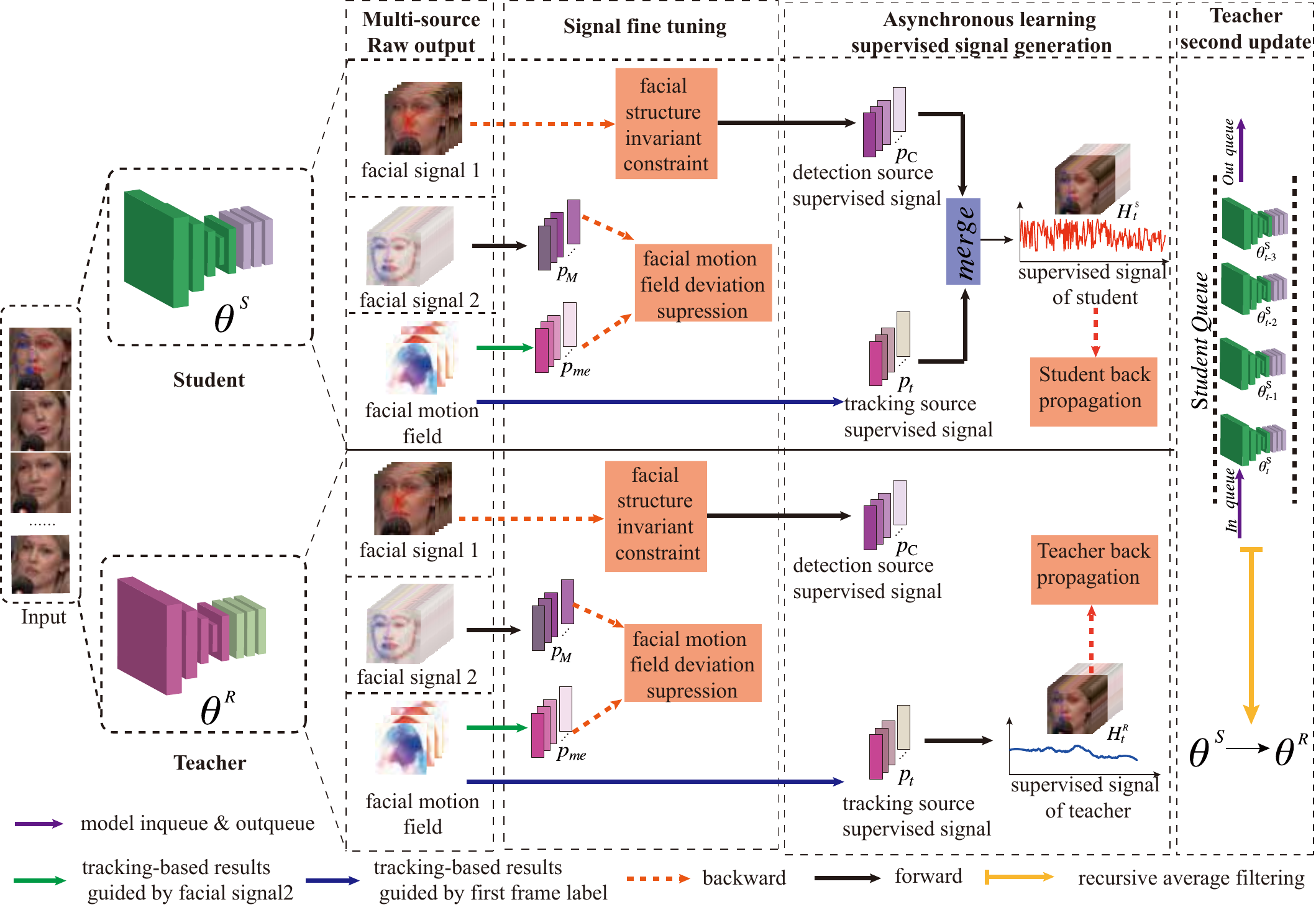} % Reduce the figure size so that it is slightly narrower than the column.
		\caption{Overview of the student and teacher's pipeline. $\theta^{S}$ and $\theta^{R}$ are parameters of student and teacher respectively. Facial signal 1 is Contour Landmark Offset Regression-based results, and Facial signal 2 is Landmark Heatmap Regression-based results. Thus $p_C$, $p_{M}$ are landmark signals from detection source. And $p_{t}$,$p_{me}$ are landmark signals from tracking source.}
		\label{fig2}
	\end{figure*}
	
    The situation has motivated researchers to leverage the semi-supervised learning paradigm by making use of both labeled and unlabeled data. Many of these methods operate by propagating labeling/supervision information to unlabeled data via pseudo-labels. For example,  \citeauthor{NIPS2019_9259} presented a self-training method that takes many iterations to predict labels for the unlabeled data by gradually increasing the number of utilized samples and to retrain the model. ~\citeauthor{NIPS2019_9259}. \citeauthor{10.1145/1529282.1529735, firstcotraing} presented a multi-model collaboration framework to use multiple complementary models to obtain high quality pseudo-labels better and to avoid optimizer from falling into local optimum~\cite{10.1145/1529282.1529735, firstcotraing}. In general, due to the noise in the unlabeled data, the performances of these approaches vary a lot depending on the way the pseudo-labels are generated and utilized.
    
    Amongst the reported methods, self-supervision signal generated by the teacher-student model~\cite{NIPS2017_6719, NIPS2019_8749} has become an effective strategy for semi-supervised learning, where the output of a student model is enforced to be consistent with a teacher model for unlabeled input. By asynchronously updating the two models, the consistency constraint helps to mine high quality pseudo-labels that can bring significant improvement in semi-supervised learning tasks.
    
    The strategy inspired us to propose a facial landmark detection model based on asynchronous-learning with consistency constraint from multi-source supervision signals. The method consists of the teacher-student model where a radical student is updated with raw multi-source and a calm teacher is updated with more stable gradient. Specifically, the sources of input include firstly a set of facial landmark targets predicted through a facial \ac{MFE} module. Secondly, we use detection method to obtain the second set of facial landmark targets through the \ac{LHR}. And thirdly, another set of facial landmark targets is obtained through face center detection and \ac{CLOR}. With the three sets of signals from different sources, the radical student uses all three types of signals to update parameters, while the calm teacher only uses facial \ac{MFE} and \ac{LHR} to update parameters. To allow the teacher model to accept part of the student's suggestions, an exponential moving average strategy is additionally used to update the teacher’s parameters again, before the teacher model instructs the student model. In this way, the overall framework can utilize the consistency constraint between sources and models to achieve satisfactory facial landmark detection performance. Fig.\ref{fig2} shows an overview of our asynchronous-learning framework. In summary, major contributions of the paper include:
    	
    \begin{itemize}
    	\item Three improved sets of supervision source to train the teacher-student network model, including the facial \ac{MFE}, the \ac{LHR} and the \ac{CLOR};
    	\item A teacher-student network model with asynchronous-learning to effectively smooth the learning and to obtain improved performance as shown in our experiments.
    \end{itemize}

	\section{Related Work}
	\textbf{Fully-supervised Facial Landmark Detection.} Fully supervised facial landmark detection can be categorized into two types: coordinate regression~\cite{Xiong_2013_CVPR, cite-key-Xudong} and landmark heatmap regression~\cite{Wei_2016_CVPR, Dong_2018_CVPR,10.1007/978-3-319-46484-8_29} according to the type of supervision signal.
	~\citeauthor{zhou2019objects} leverages the idea of "object as point" to regress anchor of object. Based on this idea, we design \ac{CLOR} task and obtained another type of facial landmark supervision signal. These different source of supervision signals are our model's operation objects.
	
	\subsubsection{Semi-supervised Facial Landmark Detection}
	Semi-supervised facial landmark detection in video aims to use less annotation data to improve the performance of the entire video. Most of the existing arts focuses on how to mine pseudo labels to expand the training set.
	
	~\citeauthor{dong2018sbr} uses a differentiable optical flow estimation method to obtain pseudo labels of subsequent frames, and contributes a way to estimate facial motion field, and completes the whole tasks through a two-stage training process, which performs pseudo label mining implicitly. When it fine-tunes the detector with tracking-based results in the second stage, it tries to solve the problem proposed by ~\citeauthor{8237671} that detection results have no drift but low accuracy.
	
	Fully supervised facial landmark detection mentioned above are for single-frame images, which are also applicable to facial landmark detection in video. Based on these fully supervised method, some self-training or co-training approaches simply leverage confidence score or an unsupervised loss to mine qualified samples. Due to the complementarity between multiple models, researchers proposed to leverage multiple models to promote each other's performance. ~\cite{hinton2015distilling,10.1007/978-3-319-73603-7_40} as classic teacher-student models aim to let student model fit teacher's output. ~\cite{Dong_2019_ICCV} contains two different networks and leverages pseudo-labels with high quality envaluated by teacher to train student. Different from these multi models, our TSAL framework consists of two networks with exactly the same structure. And we use supervised signals with different levels of disturbance from different sources to train student and teacher, and implicitly mine high quality pseudo-labels through a mechanism of asynchronously updating network parameters.
	
	\section{Methodology}
	\subsection{Motivation}
	Different from method of regressing face coordinates directly, we design a Contour Landmark Offset Regression~(\ac{CLOR}) task to detect facial landmarks $p_C$. Inspired by ~\cite{zhou2019objects}, we regard face as a point, and use a Gaussian template $Y_{xy}=e^{-\frac{(x-ctx)^2+(y-cty)^2}{{\sigma(box_w, box_h)}^2}}$ to represent face as an isotropic 2D Gaussian distribution. And we leverage another parallel branch to regress the offsets $O$ from remaining landmarks to the center directly. However, this simple yet designed task \ac{CLOR}  has a large detection variance on consecutive frames. Fortunately, $p_C$ can maintain a stable facial structure.
	
	Landmark Heatmap Regression~(\ac{LHR}) based methods~\cite{dong2018sbr, wayne2018lab} often regress a $N$-channel landmark heatmap $H$ firstly, and then parses landmark coordinates $h_{M}$ in post process, where $N$ is the number of landmarks. However, when face encounters occlusion, the response of landmarks is weakened. When we parse heatmap into coordinates in post process, the facial structure will be deformed. Therefore, we attempt to apply constraint to correct $p_C$ from \ac{CLOR}, and regard it as a kind of signal for unlabeled images to supervised $h_M$ from \ac{LHR}.
	
	\ac{LHR}-based method and \ac{CLOR}-based method are both detection-based methods in facial landmark detection. In continuous video frames, if mutual information of inter-frames is fully utilized, the facial landmark detection will be more accurate. Thus we leverage an unsupervised learning method of motion field estimation proposed in ~\cite{10.1007/978-3-030-20893-6_23} to estimate the facial motion field. However, just as ~\cite{8237671} mentioned, tracking-based method has drift although with a high accuracy. Therefore, we try to eliminate this drift by maintain the consistency between detection-based landmark detection results and tracking-based landmark detection results.
	
	The core of improving performance of detector is to make full use of unlabeled images by mining higher-quality pseudo labels to participate in training. In the whole system, we obtain two supervision signals from different sources, one comes from detection source and the other comes from tracking source. However, noise of detection source signal is obvious. In order to smooth disturbance in supervision signal, we let the two supervision signals run on two asynchronously updated models, and use recursive average filtering to filter out the noise in supervision signal. 
	
	\subsection{The Same Pipeline of Teacher and Student}
	Input of the system are $K$ frames of the video, denoted as $\{{I_{{0}}^l, I_{1}^u,...,I_{K-1}^u}\}$, and only $I_{{0}}^l$ has labele $p^{l} \in \mathbb{R}^{N \times 2}$. Teacher and student of TSAL framework have exactly the same structure, but have different back propagation. For ease of description, we indiscriminately express the same pipeline of teacher and student in this section. 
	
	We have a encoder-decoder network to perform facial \ac{LHR} and facial \ac{CLOR}. One of output of decoder is facial landmark heatmap $H \in \mathbb{R}^{K \times N \times w\times h}$, the other output is face center heatmap $C$, and another output is offset $O$ from remaining landmarks to the center. Thus we obtain two groups of facial landmark coordinates $p_C\in \mathbb{R}^{K\times  N\times 2}$ and $p_M\in \mathbb{R}^{K\times  N\times 2}$ from \ac{CLOR} and \ac{LHR} respectively. And motion field estimation module outputs inter-frame motion field estimation $V$, so the tracking-based landmark detection results are represented as $\phi(V, g)$, where $g$ is the guide~(landmark coordinates) of the first frame.
	
	Therefore, the total loss function of supervised detection and unsupervised tracking is defined as follows:
	\begin{eqnarray}
	\ell_{s} = \mathcal L_{I} +\omega(t) \mathcal L_{D} +\ell_{M}^s + \ell_{C}^s(C,O) +\ell_{E}(\{I_{(i)}\}_{i=0}^{K-1})\label{eq1},
	\end{eqnarray}
	where $\mathcal L_{I}$ is facial structure invariant constraint we proposed to alleviate large variance of \ac{CLOR}-based results, and $\mathcal L_{D}$ represents the loss of motion field deviation suppression. $\ell_{M}^s $ is multi-channel landmark heatmap regression error by ~\citeauthor{dong2018sbr}, $\ell_{C}^s(C,O)$ is landmark coordinate regression error by ~\citeauthor{zhou2019objects}, $\ell_{E}(\{I_{(i)}\}_{i=0}^{K-1})$ is facial motion estimation loss proposed in ~\cite{10.1007/978-3-030-20893-6_23}. $\omega (t)$ is a piecewise exponential climbing function helps the system gradually adapt to unsupervised signals.
	
	\begin{figure}[!t]
		\centering
		\includegraphics[width=0.9\columnwidth]{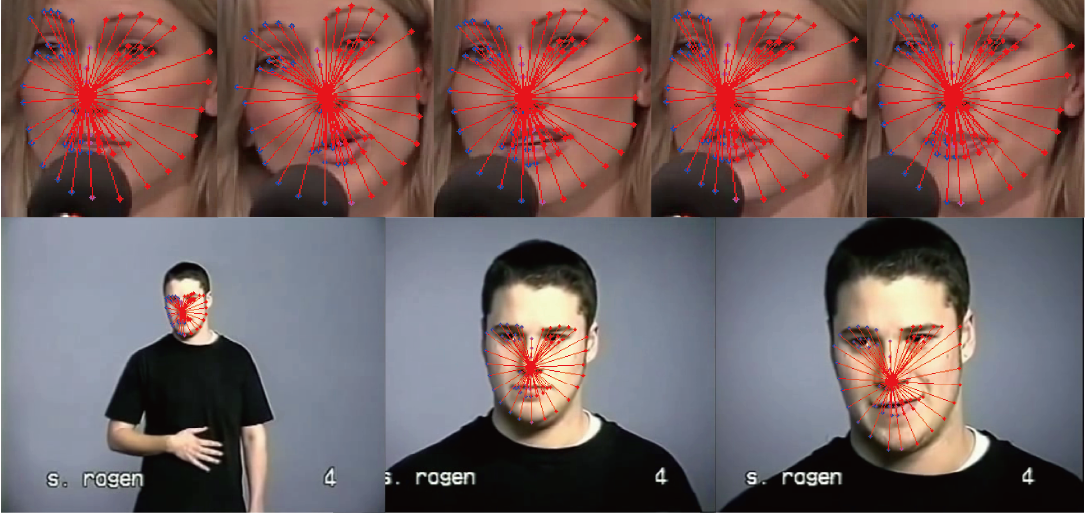} % Reduce the figure size so that it is slightly narrower than the column. Don't use precise values for figure width.This setup will avoid overfull boxes.
		%\vspace{-0.1cm}
		\caption{Facial structure invariant constraint ensures facial landmarks are always attracted by face center on the adjacent frames. And face size are used to normalize facial structure invariant constraint.}
		\label{fig3}
	\end{figure}
	\subsubsection{Facial Structure Invariant Constraint}
	Through \ac{LHR} and \ac{CLOR}, we have obtained two groups of facial landmark coordinates $p_{M}$ and $p_{C} $ respectively. When the facial motion range is too large or there is an object occlusion, $p_{M}$ has a deviation when parsing coordinates from heatmap, which deforms the facial landmark structure. While $p_{C} $ is heatmap-parsing free, so the facial landmark structure of $p_{C} $ is stable, but the variance of the $p_{C} $ in adjacent frame is large. We hope to merge $p_{M}$ and $p_{C} $ reasonably to alleviate the structure deformation of facial landmarks of $p_{M}$ in complex situations.
	
	Therefore, we assume that on continuous $K$ frames, remaining N-1 facial landmarks are always attracted by the continuously changing face center $(ctx,cty)$ (as Fig.\ref{fig3} shown), that is, normalized offset modulus sum of landmarks to the face center in the current frame is a constant, which makes $p_{C} $ to maintain the facial landmark structure while having a small variance between adjacent frames. The facial structure invariant constraint about facial landmark is defined as follows:
	\begin{eqnarray}
	\mathcal L_{I} = \frac{1}{K}\sum_{i=0}^{K-2}\sum_{j=i+1}^{K-1}(\frac{||O^{(i)}||}{box_w^i}-\frac{||O^{(j)}||}{box_w^j})^2\label{eq2},
	\end{eqnarray}
	where $box_w^i$ is width of face bounding box in i-th frame. Through coordinates $p_C^i$, we  have $box_w$ = $max(\{x_n\}_{n=1}^{N})$ - $min(\{x_n\}_{n=1}^{N})$, $box_h$ = $max(\{y_n\}_{n=1}^{N})$ - $min(\{y_n\}_{n=1}^{N})$. In final test session, the results are obtained through interpolation of $p_{M}$ and $p_{C} $,  which is $p = \gamma p_{M}+(1-\gamma) p_{C}$, and we set $\gamma = 0.8$ in our final test. 

	\subsubsection{Motion Field Deviation Suppression}
	When using $p_{M}^{(0)}$ obtained by LHR as a facial motion filed guide to generate facial landmark of subsequent frames $p_{me}$, the inaccuracy of the motion field $V$ estimated by facial motion filed estimation module will make $p_{me}^{(K-1)}$ inconsistent with detection-based results $p_{M}^{(K-1)}$, which is reason why facial landmarks of subsequent frames predicted by tracking-based method have drift. 
	
	In this case, we propose Motion field deviation suppression to maintain the consistency between facial landmark detection results obtained by tracking-based method and detection-based method. Motion field deviation suppression updates parameters of facial motion filed estimation module, and the motion field estimation deviation constraint loss is calculated by:
	\begin{eqnarray}
	\mathcal L_ {D}= || \phi(V, p_M^{(0)})^{(K-1)} -p^{(K-1)}_{M} ||^ 2_2\label{eq3}.
	\end{eqnarray}
	Specifically, we represent the same pipeline loss of student and teacher as $\ell_s^{S}$ and $\ell_s^{R}$ respectively.
	
	\subsection{Different Pipelines of Teacher and Student}
	Through the same pipeline mentioned above, student and teacher obtained two groups of fine-tuned supervision signals from different source respectively. As two types pseudo-label  for unlabeled images, they have different levels of disturbance. In order to seek a balance between two supervision signals, we use two model asynchronous learning and recursive average filtering methods to filter out the noise in the supervision signal, and effectively ensemble two supervision signals to update network parameters, which is the source of motivation for mining much more significant pseudo-labels.
	
	Outputs of motion field estimation module in teacher and student are facial motion field $V^{R}$ and $V^{S}$ respectively, then we obtain tracking-based landmark coordinates of teacher and student, which are represented as $\phi(V^{R}, p_l)$ and $\phi(V^{S}, p_l)$. The other supervision signals of teacher and student are \ac{CLOR}-based $p_C^{R}$ and $p_C^{S}$. And teacher merely uses tracking-based stable signal to update parameters, that's to say, final supervision signal for \ac{LHR}-based output $H^{R}_t$ is $\phi(V^{R}, p_l)$.
	Then the loss function of teacher using pseudo-heatmap supervision can be defined as:
	\begin{eqnarray}
	\mathcal L_ {d}^{R} = \frac{\sum_{i=1}^{K-1}|| H^{{R;(i)}} -\phi ^{(i)}(V^{R}, p_l) ||^ 2_2}{Nwh(K-1)}\label{eq4}.
	\end{eqnarray}
	
	While supervision signal of radical student is also affected by $p_C^{S}$. We use linear interpolation to merge $p_C^{S}$ and $\phi(V^{S}, p_l)$, and final supervision signal for \ac{LHR}-based output $H^{S}_t$ is $ \mu \phi(V^{S}, p_l)+(1-\mu)p_C^{S}$. Then the loss function of student using pseudo-heatmap supervision can be defined as
	\begin{eqnarray}
	\mathcal L_ {d}^{S} = \frac{\sum_{i=1}^{K-1}|| H^{S;(i)} -\phi ^{(i)}((V^{S}, p_l),p_C^{S}, \mu) ||^ 2_2}{Nwh(K-1)}\label{eq5}.
	\end{eqnarray}
	
	Therefore, from the same pipeline and different pipelines mentioned above, we express total loss of calm teacher and radical student as follows:
	\begin{eqnarray}
	\mathcal L^{R} = \ell_{s}^{R} +\omega (t) \mathcal L_{d}^{R}\label{eq6},
	\end{eqnarray}
	\begin{eqnarray}
	\mathcal L^{S} = \ell_{s}^{S} + \omega(t)  \mathcal L_{d}^{S}\label{eq7},
	\end{eqnarray}
	where $\omega (t)$ is a function of training iteration mentioned in the same pipeline above, used to control the proportion of unsupervised signals.
	
		\begin{algorithm}[t]  
		\caption{Asynchronous-learning model }  
		\KwIn{Teacher $\Theta_{t-1}$at step $t-1$ and Student queue $T[\Theta^{S}_{t-v},...,\Theta^{S}_{t-1}]$.}  
		\KwOut{Teacher $\Theta_{t}^{R}$ at step $t$. }  
		\For{$step=t;step \le N; step++$}  
		{  
			$\Theta_{t}^{R}$$\leftarrow$$\Theta_{t-1}^{R}$-$\eta$$\frac{\partial \mathcal L^{R}_{t}}{\partial \Theta_{t-1}^{R}}$
			
			$\Theta_{t}^{S}$$\leftarrow$$\Theta_{t-1}^{S}$-$\eta$$\frac{\partial \mathcal L^{S}_{t}}{\partial \Theta_{t-1}^{S}}$
			
			$T[\Theta^{S}_{t-v+1},...,\Theta^{S}_{t}]$ $\leftarrow$$T[\Theta^{S}_{t-v},\Theta^{S}_{t-v+1}...,\Theta^{S}_{t-1}]$
			
			$\Theta$ = $\textbf{0}$
			
			\For{$i=0;i < v; i++$}  
			{  
				$\Theta$ $\leftarrow$ $\Theta$+$T[i]$
			}  
			$\Theta_{t}^{R}$ $\leftarrow$ $\Theta$/$v$
		}  
		return $\Theta_{t}^{R}$\;  
	\end{algorithm}  
	
	\subsection{Parameter Updating}
	Student and teacher asynchronously update their network parameters $\Theta^{S}$ and $\Theta^{R}$ with different supervised signals from different sources in same task through $\mathcal L_{S}$ and $\mathcal L_{R}$ respectively. Since the pseudo-label supervision signal used by student with more disturbance, in order to allow teacher to accept student’s suggestions reasonably, we use a recursive average filtering method to smooth the noise in student.
	
	Specifically, we maintain a queue $T$ of length $v$ to store recent iteration student's parameters, denoted as $T[\Theta^{S}_{t-v+1}$,$\Theta^{S}_{t-v+2}$, ...,
	$\Theta^{S}_{t}]$. And we define $\Theta_t^{R}$ at training step $t$ as the teacher of successive weights:
	\begin{eqnarray}
	\Theta^{R}_t  = \lambda \Theta^{R}_{t} + (1-\lambda)Avg(T)\label{eq8},
	\end{eqnarray}
	where $\lambda$ is a smoothing coefficient hyperparameter, and $Avg$ means calculating students' parameter average value in queue $T$. ~\cite{NIPS2017_6719} is a special case when our queue length is 1. And when we set $v=1$, following ~\cite{NIPS2017_6719}, we let $\lambda$=0.999. Such a recursive average filtering method helps teachers absorb suggestions from students to promote teacher stably.

	\section{Experiments}
	\subsection{Dataset}
	\subsubsection{Image}
	The first image dataset used was the 300W dataset~\cite{Sagonas_2013_ICCV_Workshops} which is a combination of five other datasets, including the LFPW, the AFW, the HELEN, the XM2VTS, and the IBUG dataset. Following prior works, our training set included the training of LFPW, HELEN as well as the full set of AFW, in which there was 3148 images in total. The common test subset consisted of 554 test images from LFPW and HELEN, and the challenging test subset consisted of 135 images from IBUG. The full test set was the union of the common and the challenging subsets, with 689 images in total.
	
	The second image dataset used was the AFLW dataset~\cite{6130513} that consists of 25993 faces from 21997 real-world images~\cite{8099876, dong2018sbr}. Following~\citeauthor{Zhu_2016_CVPR}, the dataset was partitioned into two different subsets, AFLW-Full and AFLW-Front respectively. The two subsets have the same training set, but with different testing samples: AFLW-Full contains 4386 test samples, while AFLW-Front only uses 1165 samples from AFLW-Full as testing set.

	\subsubsection{Video}
	The video dataset used was the 300VW dataset~\cite{Shen_2015_ICCV_Workshops} that contains 50 training videos with 95192 frames. The test set consists of three (A, B and C) with 62135, 32805 and 26338 frames, respectively, and subset C is the most challenging one.  Following~\cite{8237671}, we report the results on subset C.
	
	\subsection{Setup}
	
	\subsubsection{Training}
    We selected as our training set 20 videos from the 300VW training dataset (1st, 2nd, 7th, 13th, 19th, 20th, 22th, 25th, 28th, 33th, 37th, 41th, 44th, 47th, 57th, 119th, 138th, 160th, 205th and 225th) with different brightness, scenes, facial motion amplitude, face scale, occlusion, and gender. We set a stride of 10 frames which means that only the first frame in every 10-frame sequences was annotated, and the remaining 9 frames were used as unlabeled data.
	
    \subsubsection{Metric}
    For image datasets (300W and AFLW) the \ac{NME} normalized by inter-pupil distance and face size was used respectively, while for video dataset (300VW), the mean Area Under the \ac{AUC}@0.08 error~\cite{8237671} was employed.
	
\subsubsection{Implementation}
    
The input image was resized to 256x256. We used the Adam optimizer for training with 140 epochs, with an initial learning rate of $1.25e-4$, decayed by $10$ and $100$ in the 90th and 120th epochs. The setting of the power value in Eq.~\eqref{eq1},\eqref{eq6} and \eqref{eq7} followed~\cite{NIPS2017_6719}, and the climbing period was from 1 to 60 epochs, the retention period was from 61 to 110 epochs, and the decay period was from 111 to 140 epochs. The batch size was set to 8 for both the teacher and the student model, and we used random flip, random translation, random angle rotate, and color jitter for data augmentation. All our experiments were conducted on a workstation with 2.4GHz Intel Core CPUs and 4 NVIDIA GTX 1080Ti GPUs.
    
\subsection{Comparison with \ac{SOTA}}
    
\subsubsection{Results on 300W and AFLW}
	
As shown in Table~\ref{table1}, compared with SDM~\cite{Xiong_2013_CVPR}, TCDCN~\cite{10.1007/978-3-319-10599-4_7}, CPM~\cite{Wei_2016_CVPR},LAB~\cite{wayne2018lab}, Wingloss~\cite{DBLP:journals/corr/abs-1711-06753} and PFLD1x~\cite{guo2019pfld}, our model presented clear improvement in the 300W dataset. Similarly, in the AFLW dataset, our model also outperformed SDM~\cite{Xiong_2013_CVPR}, LAB~\cite{wayne2018lab}, LBF~\cite{6909614}, CCL~\cite{Zhu_2016_CVPR}, Two stage~\cite{Lv_2017_CVPR}, SAN~\cite{dong2018style} and DSRN~\cite{Miao_2018_CVPR}.

\subsubsection{Results on 300VW}
As shown in Table\ref{table2}, compared with DGCM~\cite{8237671}, SBR$^*$~\cite{dong2018sbr} and TS$^3$~\cite{Dong_2019_ICCV}, our model achieved the best accuracy for the 300VW dataset. It is worth noting that, although TS$^3$~\cite{Dong_2019_ICCV} also employs the teacher-student architecture, with the supervision from multiple sources, our model uses a simpler structure to achieve even higher detection accuracy. 

%Compared with another teacher-student framework TS$^3$~\cite{Dong_2019_ICCV}, we have a totally different teacher-student communication mechanism. However, TS$^3$ only have a public NME result normlized by inter-ocular distance. Due to the relationship of ION and IPN, our dual-model is better than TS$^3$ in theory. As an auxiliary module, motion field estimation module is only used during training. Therefore, compared with existing work, we do not significantly increase the complexity of test model but with higher detection accuracy. 
	
	   \begin{table}[]
		\centering
		\begin{tabular}{p{1.5cm}<{\centering}|p{1.0cm}<{\centering}p{1.0cm}<{\centering}p{1.1cm}<{\centering}|p{0.6cm}<{\centering}p{0.6cm}<{\centering}}
			\hline
			\multirow{2}{*}{Method} & \multicolumn{3}{c|}{300W}     & \multicolumn{2}{c}{AFLW} \\
			& Common & Challenge & Full & Front & Full       \\ \hline
			\hline
			SDM                     & 5.57   & 15.40     & 7.52     & 2.94        & 4.05       \\
			LBF                     & 4.95      & 11.98         & 6.32        & 2.74        & 4.25       \\
			TCDCN                   & 4.80   & 8.60      & 5.54     & -           & -          \\
			LAB                     & 3.42   & 6.98      & 4.12     & 1.62        & 1.85       \\
			CPM                     &3.39    &8.14       &4.36      & -           & -          \\
			Wing loss               & 3.27   & 7.18      & 4.04     & -           & \textbf{1.65}       \\
			PFLD 1x                 & 3.32   & 6.56      & 3.95     & -           & 1.88          \\
			CCL                     & -      & -         & -        & 2.17        & 2.72       \\
			Two stage               & -      & -         & -        & -           & 2.17       \\
			SAN                     & -      & -         & -        & 1.85        & 1.91       \\
			DSRN                    & -      & -         & -        & -           & 1.86       \\
			\hline
			\textbf{Ours} & \textbf{3.13}   & \textbf{6.02}      & \textbf{3.69}     & \textbf{1.47}        & \textbf{1.65}       \\ \hline
		\end{tabular}
		\caption{ Comparison of NME with the state-of-the-art methods on 300W and AFLW datasets.}
		\label{table1}
	\end{table}
	
	\begin{table}[!t]
		\centering
		\begin{tabular*}{\hsize}{@{}@{\extracolsep{\fill}}c|cccc@{}}
			\hline
			Method              & DGCM  & SBR* & TS$^{3}$   & \textbf{Ours}  \\ \hline
			\hline
			AUC@0.08 error & 59.38 & 59.39       & 59.65 & \textbf{59.92} \\ \hline
		\end{tabular*}
		\caption{Comparisons of mean AUC@0.08 error with the state-of-the-art methods on 300VW dataset.}
		\label{table2}
	\end{table}
	
\subsection{Ablation Study}

To understand the effectiveness of each key components in our model, including the face invariance constraint, the facial motion field deviation suppression, and the dual-model asynchronous learning strategy, we conducted several ablation experiments as reported in Table~\ref{table3}.

% \usepackage{multirow}
\begin{table*}[!htp]
	\centering
	\begin{tabular}{p{2cm}<{\centering}|p{0.8cm}<{\centering} p{1.2cm}<{\centering}p{2.0cm}<{\centering}|p{1.5cm}<{\centering} p{1.5cm}<{\centering}p{1.5cm}<{\centering}|p{0.8cm}<{\centering} p{0.8cm}<{\centering}|p{1.5cm}<{\centering}}
		\hline
		\multirow{3}{*}{Model}        & \multirow{3}{*}{SIC}                & \multirow{3}{*}{MFDS}           & \multirow{3}{*}{\begin{tabular}[c]{@{}c@{}}Asynchronous\\ learning\end{tabular}}         & \multicolumn{6}{c}{Metric}                                                            \\ \cline{5-10}
		&                                     &                                 &                           & \multicolumn{3}{c|}{300W} & \multicolumn{2}{c|}{AFLW}         & \multirow{2}{*}{300VW} \\ 
		&                                     &                                 &                                    & Common & Challenge & Full & Front            & Full           &                        \\ \hline
		\hline
		baseline                      &$\times$                             &$\times$                          &$\times$                            & 3.73   & 6.92      & 4.35 &1.98              &1.66            & 55.14                  \\ \hline
		\multirow{3}{*}{single model} &\checkmark                           &$\times$                         &$\times$                            & 3.50   & 6.63      & 4.11 &1.81              &1.57            & 56.76                  \\
		&$\times$                             &\checkmark                       &$\times$                            & 3.58   & 6.72      & 4.19 &1.84              &1.57            & 55.79                  \\
		&\checkmark                           &\checkmark                       &$\times$                            & 3.39   & 6.44      & 3.98 &1.76              &1.53            & 57.58                  \\ \hline
		Student                       & \multirow{2}{*}{$\times$}           & \multirow{2}{*}{$\times$}       & \multirow{2}{*}{\checkmark}        & 3.62   & 6.67      & 4.21 &1.95              &1.62            & 56.21                  \\
		Teacher                       &                                     &                                 &                                    & 3.51   & 6.52      & 4.09 &1.84              &1.57            & 57.44                  \\ \hline
		Student                       & \multirow{2}{*}{\checkmark}         & \multirow{2}{*}{$\times$}       & \multirow{2}{*}{\checkmark}        & 3.36   & 6.28      & 3.93 &1.78              &1.50            & 57.97                  \\
		Teacher                       &                                     &                                 &                                    & 3.23   & 6.19      & 3.80 &1.66              &1.46            & 57.12                  \\ \hline
		Student                       & \multirow{2}{*}{$\times$}           & \multirow{2}{*}{\checkmark}     & \multirow{2}{*}{\checkmark}        & 3.51   & 6.45      & 4.08 &1.83              &1.51            & 56.94                  \\ 
		Teacher                       &                                     &                                 &                                    & 3.34   & 6.32      & 3.92 &1.71              &1.49            & 58.11                  \\ \hline
		Student                       & \multirow{2}{*}{\checkmark}         & \multirow{2}{*}{\checkmark}     & \multirow{2}{*}{\checkmark}        & 3.24   & 6.12      & 3.79 &1.73              &1.50            & 58.89                  \\ 
		Teacher                       &                                     &                                 &                                    & \textbf{3.13}   & \textbf{6.02}      & \textbf{3.69} &\textbf{1.65}              &\textbf{1.47}            & \textbf{59.92}                  \\ \hline
	\end{tabular}
	\caption{\ac{NME} scores with respect to on/off between the facial structure invariant constraint~(SIC), the motion field deviation suppression~(MFDS) and the teacher-student asynchronous-learning strategy on 300W, AFLW and 300VW datasets.}
	\label{table3}
\end{table*}

%prove the effectiveness of the face invariance constraint and facial motion field deviation suppression we proposed, we conducted ablation experiments in a single model and a dual model framework. 

%And we conduct further experiments to verify the effectiveness of the dual-model asynchronous update framework based on recursive average filtering on 300W, AFLW and 300VW. 

%In addition, to verify the effectiveness of the pseudo-label mined by facial motion field deviation suppression, we quantitatively analyzed the AUC@0.08 error of pseudo-label obtained by facial \ac{MFE} in the 20 unlabeled videos used in training.

\subsubsection{The Facial Structure Invariant Constraint:}

Compared with the baseline model accuracy in Table~\ref{table3}, when the SIC module was on in either the single model or the student-teacher dual-model, a clear performance gain can be observed, which shows that by enforcing the facial structure to be consistent in consecutive frames, SIC made the \ac{CLOR}-based detection results more stable and accurate.

\subsubsection{The Facial Motion Field Deviation Suppression:}

It can also be seen from Table\ref{table3} that, by adding MFDS to the \ac{MFE} module, the \ac{NME} score on 300W and the mean AUC@0.08 error rate on 300VW were improved. Therefore, the use of \ac{LHR}-based detection results to supervise the training of facial \ac{MFE} can decrease the deviation of motion estimation. The quantitative evaluation of pseudo-label quality in Fig.\ref{fig5} also verifies the effectiveness of field deviation suppression.

%In the quantitative evaluation of pseudo-label quality in Fig.\ref{fig4}, motion field deviation suppression has significantly improved the quality of pseudo-labels, which also indicate that using detection results from detection-based model to supervised tracking-based results can make motion estimation converge to a better solution, which proves effectiveness of otion field deviation suppression.
	
\subsubsection{Asynchronous-learning Strategy}

In Table~\ref{table3}, the performance of the teacher model was always better than the student model. This indicates that there was indeed a disturbance in updating the student model after merging the \ac{CLOR}-based signal and tracking-based signal, and the applied recursive average filtering did improved the quality of teacher model than only using tracking-based signal.

To summarize, it is revealed that: 1) Facial structure invariant constraint is benefit to burst landmark detection results from \ac{CLOR}; 2) Motion field deviation suppression successfully maintains the consistency between tracking-based detection results and detection-based detection results, alleviating the problem of field drift; And 3) Dual-model asynchronous learning and recursive filtering method helped mining high-quality pseudo-labels.

\subsection{Qualitative Analysis}
	
To further verify the effectiveness of facial invariant constraint and motion field deviation suppression, we visualize facial landmarks  detection results of several faces on 300W test datasets in Fig.\ref{fig5}. Compared with baseline, we can see that facial invariant constraint and motion field deviation suppression make the contour of facial landmarks more compact, which can effectively maintain the facial landmark structure invariant and improve the detection accuracy of landmarks. Combining facial structure invariant constraint and motion field deviation suppression benefits to reduce drift of facial motion estimation.
	
	\begin{figure}[!t]
		\centering
		\includegraphics[width=0.9\columnwidth]{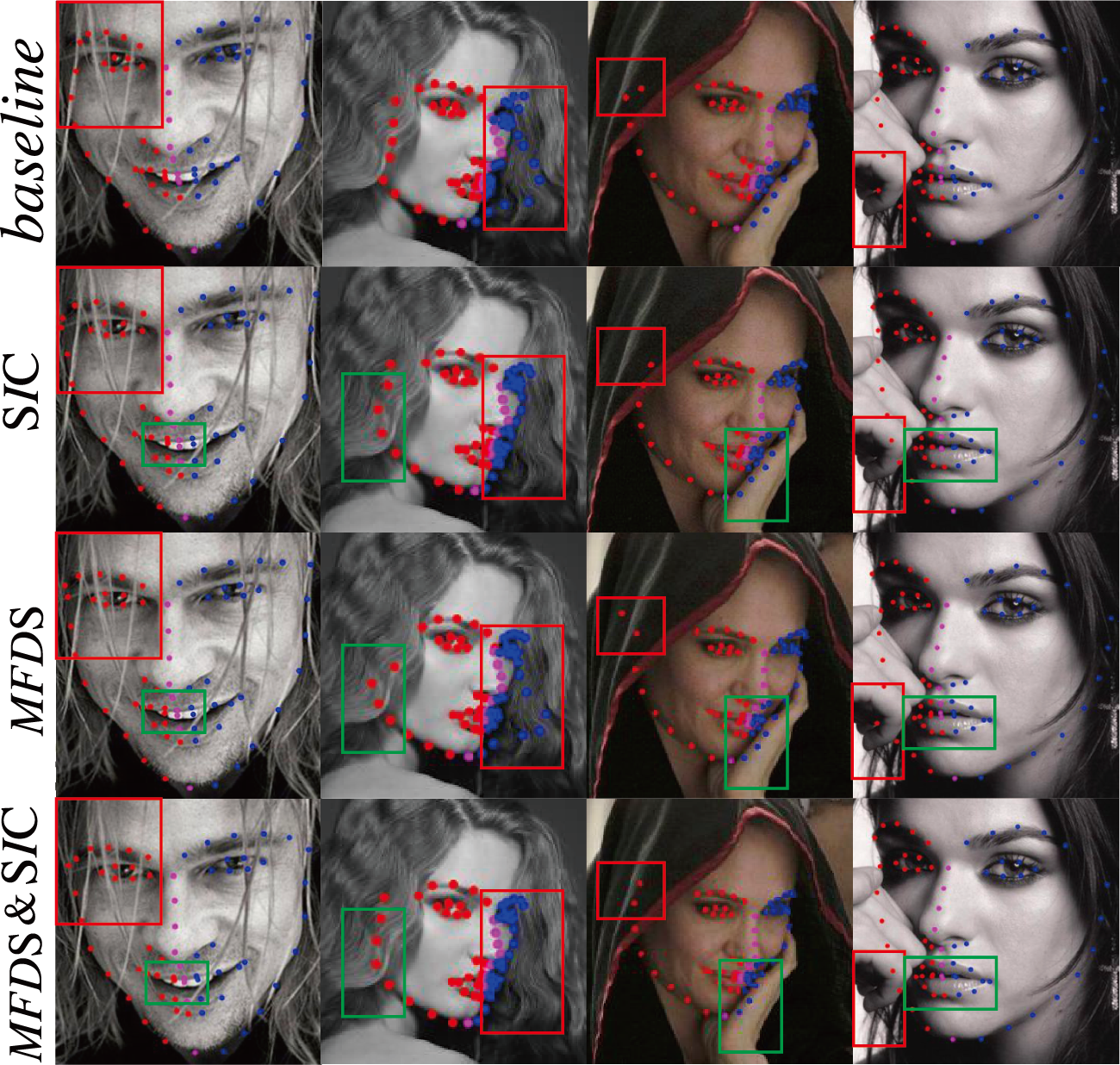} % Reduce the figure size so that it is slightly narrower than the column. Don't use precise values for figure width.This setup will avoid overfull boxes.
		%\vspace{-0.1cm}
		\caption{Qualitative results on several faces in 300W challenge dataset by teacher of our model.}
		\label{fig4}
	\end{figure}
	
	\begin{figure*}[!t]
	\centering
	\includegraphics[width=1.0\textwidth]{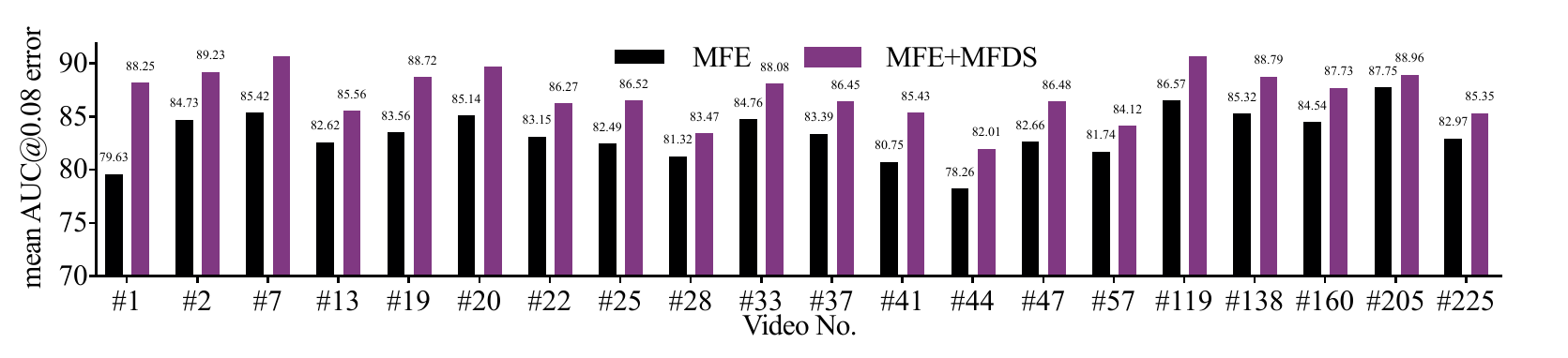} % Reduce the figure size so that it is slightly narrower than the column.
	%\vspace{-0.5cm}
	\caption{Mean AUC@0.08 error of pseudo-labels estimated by motion field estimation~(MFE) module on the unlabeled training samples in the 20 videos used in training.}
	\label{fig5}
	\end{figure*}
	
	\begin{figure*}[!t]
	\centering
	\includegraphics[width=1.0\textwidth]{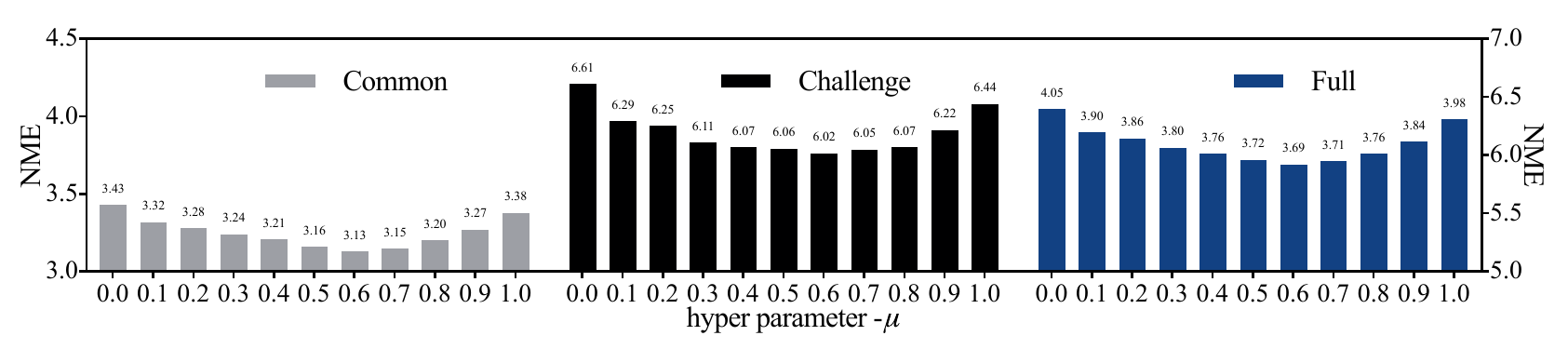} % Reduce the figure size so that it is slightly narrower than the column.
    %\vspace{-0.5cm}
	\caption{Analysis of hyper-parameter $\mu$ on 300W test datasets.}
	\label{fig6}
	\end{figure*}

\subsection{Discussion}

	During training, we have a hyper parameter $\mu$ in Eq.\ref{eq5} that needs to be pre-defined. $\mu$ is used to control the fusion of tracking-based and \ac{CLOR}-based detection results to generate multi-source supervision signals of student. Fig.\ref{fig6} shows the NME of 300W test datasets with different $\mu$ under the teacher-student asynchronous learning framework, using the facial SIC to correct \ac{CLOR}-based detection results, and using MFDS to correct tracking-based detection results, from which we can find that:~\textbf{(1)} The worst result comes from $\mu$=0. At this time, the student’s supervision signal comes entirely from \ac{CLOR}-based detection results, which means that there is a large disturbance in the \ac{CLOR}-based detection results. ~\textbf{(2)} The best result comes from  $\mu$=0.6. At this time, the best balance is achieved between \ac{CLOR}-based and tracking-based detection results. ~\textbf{(3)} When A is too large, NME starts to increase instead, which indicates that the unity of teacher-student supervision signals in the asynchronous learning framework will inhibit the promotion of teacher.
	
\section{Conclusion}

	In this paper, we propose a teacher-student asynchronous learning framework for facial landmark detection, which shows effectiveness when mining pseudo-labels of unlabeled face video frames. Additionally, we propose a facial structure invariant constraint to fine tune Contour Landmark Offset Regression-based coordinates and a motion field deviation suppression method to maintain the consistency between detection-based and tracking-based landmark coordinates, which improves performance of our model significant.

	\bigskip

	\bibliography{Bibliography-File}

\begin{thebibliography}{40}
\providecommand{\natexlab}[1]{#1}
\providecommand{\url}[1]{\texttt{#1}}
\providecommand{\urlprefix}{URL }
\expandafter\ifx\csname urlstyle\endcsname\relax
  \providecommand{\doi}[1]{doi:\discretionary{}{}{}#1}\else
  \providecommand{\doi}{doi:\discretionary{}{}{}\begingroup
  \urlstyle{rm}\Url}\fi

\bibitem[{Berthelot et~al.(2019)Berthelot, Carlini, Goodfellow, Papernot,
  Oliver, and Raffel}]{NIPS2019_8749}
Berthelot, D.; Carlini, N.; Goodfellow, I.; Papernot, N.; Oliver, A.; and
  Raffel, C.~A. 2019.
\newblock MixMatch: A Holistic Approach to Semi-Supervised Learning.
\newblock In \emph{NIPS 32}, 5049--5059.

\bibitem[{{Blanz} and {Vetter}(2003)}]{1227983}
{Blanz}, V.; and {Vetter}, T. 2003.
\newblock Face recognition based on fitting a 3D morphable model.
\newblock \emph{IEEE Transactions on Pattern Analysis and Machine Intelligence}
  25(9): 1063--1074.

\bibitem[{Blum and Mitchell(1998)}]{firstcotraing}
Blum; and Mitchell. 1998.
\newblock Combining Labeled and Unlabeled Data with Co-Training.
\newblock New York, NY.

\bibitem[{Cao et~al.(2014)Cao, Wei, Wen, and Sun}]{cite-key-Xudong}
Cao, X.; Wei, Y.; Wen, F.; and Sun, J. 2014.
\newblock Face Alignment by Explicit Shape Regression.
\newblock \emph{International Journal of Computer Vision} 107(2): 177--190.

\bibitem[{Dong et~al.(2018{\natexlab{a}})Dong, Yan, Ouyang, and
  Yang}]{Dong_2018_CVPR}
Dong, X.; Yan, Y.; Ouyang, W.; and Yang, Y. 2018{\natexlab{a}}.
\newblock Style Aggregated Network for Facial Landmark Detection.
\newblock In \emph{Proceedings of CVPR}.

\bibitem[{Dong et~al.(2018{\natexlab{b}})Dong, Yan, Ouyang, and
  Yang}]{dong2018style}
Dong, X.; Yan, Y.; Ouyang, W.; and Yang, Y. 2018{\natexlab{b}}.
\newblock Style Aggregated Network for Facial Landmark Detection.

\bibitem[{Dong and Yang(2019)}]{Dong_2019_ICCV}
Dong, X.; and Yang, Y. 2019.
\newblock Teacher Supervises Students How to Learn From Partially Labeled
  Images for Facial Landmark Detection.
\newblock In \emph{Proceedings of the IEEE/CVF International Conference on
  Computer Vision (ICCV)}.

\bibitem[{Dong et~al.(2018{\natexlab{c}})Dong, Yu, Weng, Wei, Yang, and
  Sheikh}]{dong2018sbr}
Dong, X.; Yu, S.-I.; Weng, X.; Wei, S.-E.; Yang, Y.; and Sheikh, Y.
  2018{\natexlab{c}}.
\newblock {Supervision-by-Registration}: An Unsupervised Approach to Improve
  the Precision of Facial Landmark Detectors.
\newblock In \emph{Proceedings of CVPR}, 360--368.

\bibitem[{Dou, Shah, and Kakadiaris(2017)}]{Dou_2017_CVPR}
Dou, P.; Shah, S.~K.; and Kakadiaris, I.~A. 2017.
\newblock End-To-End 3D Face Reconstruction With Deep Neural Networks.
\newblock In \emph{Proceedings of CVPR}.

\bibitem[{Fabian Benitez-Quiroz, Srinivasan, and
  Martinez(2016)}]{Benitez-Quiroz_2016_CVPR}
Fabian Benitez-Quiroz, C.; Srinivasan, R.; and Martinez, A.~M. 2016.
\newblock EmotioNet: An Accurate, Real-Time Algorithm for the Automatic
  Annotation of a Million Facial Expressions in the Wild.
\newblock In \emph{Proceedings of CVPR}.

\bibitem[{Feng et~al.(2017)Feng, Kittler, Awais, Huber, and
  Wu}]{DBLP:journals/corr/abs-1711-06753}
Feng, Z.; Kittler, J.; Awais, M.; Huber, P.; and Wu, X. 2017.
\newblock Wing Loss for Robust Facial Landmark Localisation with Convolutional
  Neural Networks.
\newblock \emph{CoRR} abs/1711.06753.
\newblock \urlprefix\url{http://arxiv.org/abs/1711.06753}.

\bibitem[{Guo et~al.(2019)Guo, Li, Yu, Zhang, Ma, Ma, Liu, and
  Ling}]{guo2019pfld}
Guo, X.; Li, S.; Yu, J.; Zhang, J.; Ma, J.; Ma, L.; Liu, W.; and Ling, H. 2019.
\newblock PFLD: A Practical Facial Landmark Detector.

\bibitem[{Hinton, Vinyals, and Dean(2015)}]{hinton2015distilling}
Hinton, G.; Vinyals, O.; and Dean, J. 2015.
\newblock Distilling the Knowledge in a Neural Network.

\bibitem[{Hu et~al.(2017)Hu, Yan, Kittler, Christmas, Chan, Feng, and
  Huber}]{HU2017366}
Hu, G.; Yan, F.; Kittler, J.; Christmas, W.; Chan, C.~H.; Feng, Z.; and Huber,
  P. 2017.
\newblock Efficient 3D morphable face model fitting.
\newblock \emph{Pattern Recognition} 67: 366 -- 379.

\bibitem[{Jeong et~al.(2019)Jeong, Lee, Kim, and Kwak}]{NIPS2019_9259}
Jeong, J.; Lee, S.; Kim, J.; and Kwak, N. 2019.
\newblock Consistency-based Semi-supervised Learning for Object detection.
\newblock In \emph{NIPS 32}, 10759--10768.

\bibitem[{{Khan}, {McDonagh}, and {Tzimiropoulos}(2017)}]{8237671}
{Khan}, M.~H.; {McDonagh}, J.; and {Tzimiropoulos}, G. 2017.
\newblock Synergy between Face Alignment and Tracking via Discriminative Global
  Consensus Optimization.
\newblock In \emph{2017 ICCV}, 3811--3819.

\bibitem[{Kittler et~al.(2016)Kittler, Huber, Feng, Hu, and
  Christmas}]{cite-key}
Kittler, J.; Huber, P.; Feng, Z.-H.; Hu, G.; and Christmas, W. 2016.
\newblock 3D Morphable Face Models and Their Applications.
\newblock In Perales, F.~J.; and Kittler, J., eds., \emph{Articulated Motion
  and Deformable Objects}, 185--206.

\bibitem[{{Köstinger} et~al.(2011){Köstinger}, {Wohlhart}, {Roth}, and
  {Bischof}}]{6130513}
{Köstinger}, M.; {Wohlhart}, P.; {Roth}, P.~M.; and {Bischof}, H. 2011.
\newblock Annotated Facial Landmarks in the Wild: A large-scale, real-world
  database for facial landmark localization.
\newblock In \emph{2011 ICCV Workshops}, 2144--2151.

\bibitem[{Lee et~al.(2018)Lee, Baddar, Kim, Kim, and
  Ro}]{10.1007/978-3-319-73603-7_40}
Lee, H.~J.; Baddar, W.~J.; Kim, H.~G.; Kim, S.~T.; and Ro, Y.~M. 2018.
\newblock Teacher and Student Joint Learning for Compact Facial Landmark
  Detection Network.
\newblock In Schoeffmann, K.; Chalidabhongse, T.~H.; Ngo, C.~W.; Aramvith, S.;
  O'Connor, N.~E.; Ho, Y.-S.; Gabbouj, M.; and Elgammal, A., eds.,
  \emph{MultiMedia Modeling}, 493--504.

\bibitem[{Li, Deng, and Du(2017)}]{Li_2017_CVPR}
Li, S.; Deng, W.; and Du, J. 2017.
\newblock Reliable Crowdsourcing and Deep Locality-Preserving Learning for
  Expression Recognition in the Wild.
\newblock In \emph{Proceedings of CVPR}.

\bibitem[{Liu et~al.(2017)Liu, Wen, Yu, Li, Raj, and Song}]{Liu_2017_CVPR}
Liu, W.; Wen, Y.; Yu, Z.; Li, M.; Raj, B.; and Song, L. 2017.
\newblock SphereFace: Deep Hypersphere Embedding for Face Recognition.
\newblock In \emph{Proceedings of CVPR}.

\bibitem[{{Lv} et~al.(2017){Lv}, {Shao}, {Xing}, {Cheng}, and {Zhou}}]{8099876}
{Lv}, J.; {Shao}, X.; {Xing}, J.; {Cheng}, C.; and {Zhou}, X. 2017.
\newblock A Deep Regression Architecture with Two-Stage Re-initialization for
  High Performance Facial Landmark Detection.
\newblock In \emph{2017 IEEE Conference on Computer Vision and Pattern
  Recognition (CVPR)}, 3691--3700.

\bibitem[{Lv et~al.(2017)Lv, Shao, Xing, Cheng, and Zhou}]{Lv_2017_CVPR}
Lv, J.; Shao, X.; Xing, J.; Cheng, C.; and Zhou, X. 2017.
\newblock A Deep Regression Architecture With Two-Stage Re-Initialization for
  High Performance Facial Landmark Detection.
\newblock In \emph{Proceedings of CVPR}.

\bibitem[{Mao et~al.(2009)Mao, Lee, Parikh, Chen, and
  Huang}]{10.1145/1529282.1529735}
Mao, C.-H.; Lee, H.-M.; Parikh, D.; Chen, T.; and Huang, S.-Y. 2009.
\newblock Semi-Supervised Co-Training and Active Learning Based Approach for
  Multi-View Intrusion Detection.
\newblock In \emph{Proceedings of the 2009 ACM Symposium on Applied Computing},
  2042–2048.

\bibitem[{Miao et~al.(2018)Miao, Zhen, Liu, Deng, Athitsos, and
  Huang}]{Miao_2018_CVPR}
Miao, X.; Zhen, X.; Liu, X.; Deng, C.; Athitsos, V.; and Huang, H. 2018.
\newblock Direct Shape Regression Networks for End-to-End Face Alignment.
\newblock In \emph{Proceedings of the IEEE Conference on Computer Vision and
  Pattern Recognition (CVPR)}.

\bibitem[{Newell, Yang, and Deng(2016)}]{10.1007/978-3-319-46484-8_29}
Newell, A.; Yang, K.; and Deng, J. 2016.
\newblock Stacked Hourglass Networks for Human Pose Estimation.
\newblock In Leibe, B.; Matas, J.; Sebe, N.; and Welling, M., eds., \emph{ECCV
  2016}, 483--499.

\bibitem[{{Ren} et~al.(2014){Ren}, {Cao}, {Wei}, and {Sun}}]{6909614}
{Ren}, S.; {Cao}, X.; {Wei}, Y.; and {Sun}, J. 2014.
\newblock Face Alignment at 3000 FPS via Regressing Local Binary Features.
\newblock In \emph{2014 IEEE Conference on Computer Vision and Pattern
  Recognition}, 1685--1692.

\bibitem[{Sagonas et~al.(2013)Sagonas, Tzimiropoulos, Zafeiriou, and
  Pantic}]{Sagonas_2013_ICCV_Workshops}
Sagonas, C.; Tzimiropoulos, G.; Zafeiriou, S.; and Pantic, M. 2013.
\newblock 300 Faces in-the-Wild Challenge: The First Facial Landmark
  Localization Challenge.
\newblock In \emph{ICCV}.

\bibitem[{Shen et~al.(2015)Shen, Zafeiriou, Chrysos, Kossaifi, Tzimiropoulos,
  and Pantic}]{Shen_2015_ICCV_Workshops}
Shen, J.; Zafeiriou, S.; Chrysos, G.~G.; Kossaifi, J.; Tzimiropoulos, G.; and
  Pantic, M. 2015.
\newblock The First Facial Landmark Tracking In-the-Wild Challenge: Benchmark
  and Results.
\newblock In \emph{ICCV Workshops}.

\bibitem[{Sun, Wang, and Tang(2013)}]{Sun_2013_ICCV}
Sun, Y.; Wang, X.; and Tang, X. 2013.
\newblock Hybrid Deep Learning for Face Verification.
\newblock In \emph{ICCV}.

\bibitem[{Tarvainen and Valpola(2017)}]{NIPS2017_6719}
Tarvainen, A.; and Valpola, H. 2017.
\newblock Mean teachers are better role models: Weight-averaged consistency
  targets improve semi-supervised deep learning results.
\newblock In \emph{NIPS 30}, 1195--1204.

\bibitem[{Thies et~al.(2016)Thies, Zollhofer, Stamminger, Theobalt, and
  Niessner}]{Thies_2016_CVPR}
Thies, J.; Zollhofer, M.; Stamminger, M.; Theobalt, C.; and Niessner, M. 2016.
\newblock Face2Face: Real-Time Face Capture and Reenactment of RGB Videos.
\newblock In \emph{Proceedings of CVPR}.

\bibitem[{Tzimiropoulos(2015)}]{Tzimiropoulos_2015_CVPR}
Tzimiropoulos, G. 2015.
\newblock Project-Out Cascaded Regression With an Application to Face
  Alignment.
\newblock In \emph{Proceedings of CVPR}.

\bibitem[{Wei et~al.(2016)Wei, Ramakrishna, Kanade, and Sheikh}]{Wei_2016_CVPR}
Wei, S.-E.; Ramakrishna, V.; Kanade, T.; and Sheikh, Y. 2016.
\newblock Convolutional Pose Machines.
\newblock In \emph{Proceedings of CVPR}.

\bibitem[{Wu et~al.(2018)Wu, Qian, Yang, Wang, Cai, and Zhou}]{wayne2018lab}
Wu, W.; Qian, C.; Yang, S.; Wang, Q.; Cai, Y.; and Zhou, Q. 2018.
\newblock Look at Boundary: A Boundary-Aware Face Alignment Algorithm.
\newblock In \emph{CVPR}.

\bibitem[{Xiong and De~la Torre(2013)}]{Xiong_2013_CVPR}
Xiong, X.; and De~la Torre, F. 2013.
\newblock Supervised Descent Method and Its Applications to Face Alignment.
\newblock In \emph{Proceedings of CVPR}.

\bibitem[{Zhang et~al.(2014)Zhang, Luo, Loy, and
  Tang}]{10.1007/978-3-319-10599-4_7}
Zhang, Z.; Luo, P.; Loy, C.~C.; and Tang, X. 2014.
\newblock Facial Landmark Detection by Deep Multi-task Learning.
\newblock In Fleet, D.; Pajdla, T.; Schiele, B.; and Tuytelaars, T., eds.,
  \emph{ECCV 2014}, 94--108.

\bibitem[{Zhou, Wang, and Kr{\"a}henb{\"u}hl(2019)}]{zhou2019objects}
Zhou, X.; Wang, D.; and Kr{\"a}henb{\"u}hl, P. 2019.
\newblock Objects as Points.
\newblock In \emph{arXiv preprint arXiv:1904.07850}.

\bibitem[{Zhu et~al.(2016)Zhu, Li, Loy, and Tang}]{Zhu_2016_CVPR}
Zhu, S.; Li, C.; Loy, C.-C.; and Tang, X. 2016.
\newblock Unconstrained Face Alignment via Cascaded Compositional Learning.
\newblock In \emph{Proceedings of the IEEE Conference on Computer Vision and
  Pattern Recognition (CVPR)}.

\bibitem[{Zhu et~al.(2019)Zhu, Lan, Newsam, and
  Hauptmann}]{10.1007/978-3-030-20893-6_23}
Zhu, Y.; Lan, Z.; Newsam, S.; and Hauptmann, A. 2019.
\newblock Hidden Two-Stream Convolutional Networks for Action Recognition.
\newblock In Jawahar, C.~V.; Li, H.; Mori, G.; and Schindler, K., eds.,
  \emph{Computer Vision -- ACCV 2018}, 363--378.

\end{thebibliography}
	
\end{document}